\documentclass{article}
\usepackage{PRIMEarxiv}
\usepackage[T1]{fontenc}
\usepackage[utf8]{inputenc}
\usepackage{lmodern}
\usepackage{microtype}
\usepackage{hyperref}
\usepackage{graphicx}
\usepackage[dvipsnames]{xcolor}
\usepackage{booktabs}
\usepackage{amsmath, amssymb}
\usepackage{enumitem}
\usepackage{csquotes}
\usepackage{url}

\usepackage{tikz}
\usetikzlibrary{arrows.meta, positioning, fit, shapes, backgrounds}
\usepackage{tabularx}
\definecolor{ADSLink}{RGB}{25,25,112} 
\definecolor{ADSCite}{RGB}{34,139,34} 
\definecolor{ADSURL}{RGB}{0,51,153}   
\hypersetup{
	colorlinks=true,
	linkcolor=ADSLink,
	citecolor=ADSCite,
	urlcolor=ADSURL,
	pdfauthor={Luca Muscariello},
	pdftitle={The AGNTCY Agent Directory Service: Architecture and Implementation}
}

\title{The AGNTCY Agent Directory Service:\\Architecture and Implementation}

\author{%
	Luca Muscariello\\Cisco Systems\\\texttt{lumuscar@cisco.com}%
	\and Vijoy Pandey \\ Cisco Systems \\ \texttt{vijoy@cisco.com}%
	\and Ramiz Polic\\Cisco Systems\\\texttt{rpolic@cisco.com}%
}
\date{September 2025}

\begin{document}

\maketitle

\begin{abstract}
The Agent Directory Service (ADS) is a distributed directory for the discovery
of AI agent capabilities, metadata, and provenance. It leverages
content-addressed storage, hierarchical taxonomies, and cryptographic signing to
enable efficient, verifiable, and multi-dimensional discovery across
heterogeneous Multi-Agent Systems (MAS). Built on the Open Agentic Schema
Framework (OASF), ADS decouples capability indexing from content location
through a two-level mapping realized over a Kademlia-based Distributed Hash
Table (DHT). It reuses mature OCI / ORAS infrastructure for artifact
distribution, integrates Sigstore for provenance, and supports schema-driven
extensibility for emerging agent modalities (LLM prompt agents, MCP servers,
A2A-enabled components). This paper formalizes the architectural model,
describes storage and discovery layers, explains security and performance
properties, and positions ADS within the broader landscape of emerging agent
registry and interoperability initiatives.
\end{abstract}


\section{Introduction}
Multi-Agent Systems (MAS) increasingly rely on composable AI agents that
encapsulate specialized capabilities such as natural language processing,
reasoning, computer vision, domain analytics, orchestration, or tool mediation.
As ecosystems diversify, discovery shifts from ad-hoc curation to infrastructure
that can answer questions such as: \textit{``Which agents can perform regulated
clinical summarization with streaming JSON output?''} or \textit{``Which
reasoning agents have signed provenance and acceptable latency profiles?''}

The AGNTCY Agent Directory Service (ADS) addresses this by providing: (i) a
structured, extensible schema layer (OASF) for expressing agent capabilities,
operational traits, and evaluation signals; (ii) a distributed storage and
indexing substrate reusing OCI registry semantics for durability and
interoperability; (iii) a taxonomy-driven discovery model enabling hierarchical
capability queries; and (iv) cryptographic integrity, provenance, and
verifiability via content addressing and Sigstore-based signing.

Rather than prescribing a fixed agent execution architecture, ADS treats agent
definitions as immutable, content-addressed records enriched by extensions
(e.g., MCP server descriptors, evaluation metrics, prompt templates, feature
flags). This design maintains flexibility while supporting convergence toward
standard representational forms. Capability-based search is accelerated by a
two-level mapping: skills (and other taxonomic dimensions) to Content
Identifiers (CIDs), and CIDs to storage peers. This separation permits
horizontal scaling, adaptive replication, and low-latency intersection queries
across multiple classification axes.

The reference specification \cite{agntcy_ads_draft} and implementation
(\url{https://github.com/agntcy/dir}) emphasize reuse of proven components: OCI
distribution for artifact transport, Kademlia DHT for routing
\cite{ipfs_kademlia}, and Sigstore for cryptographic provenance. ADS complements
parallel efforts in agent registries, identity, and interoperability (e.g., MCP
\cite{modelcontextprotocolregistry}, A2A \cite{a2aspecification}, and
decentralized naming discussions harkening back to Zooko's Triangle
\cite{zooko_triangle}). In parallel, the NANDA suite of indexing and resolution
proposals ~\cite{raskar2025dnsunlockinginternetai}
~\cite{nanda2025upgrade_or_switch_arxiv} ~\cite{nanda2025adaptive_resolver}
~\cite{nanda2025enterprise_perspective} ~\cite{nanda2025registry_survey}
explores DNS-inspired, adaptive resolver, enterprise governance, and comparative
survey dimensions of an Internet-scale agent naming fabric; ADS focuses instead
on capability-centric, content-addressed discovery and can interoperate with
such name resolution layers.

\paragraph{Contributions.} This paper:
\begin{enumerate}[label=\textbf{C\arabic*}]
	\item Defines ADS architectural principles and layered model (schema,
	storage, indexing, distribution, security).
	\item Details the two-level discovery mapping and taxonomy strategy for
	scalable capability search.
	\item Explains integration of OCI / ORAS for artifact organization and
	transport.
	\item Formalizes security properties: integrity, provenance, trust
	boundaries, and threat mitigation.
	\item Presents performance and replication strategies (on-demand retrieval,
	proactive caching, strategic replication).
	\item Situates ADS among related interoperability and registry initiatives.
\end{enumerate}

\paragraph{Organization.} Section~\ref{sec:design-goals} states design goals.
Section~\ref{sec:architecture} overviews the architecture.
Section~\ref{sec:storage} covers content-addressed storage and ORAS integration.
Section~\ref{sec:discovery} details discovery, taxonomies, and DHT mappings.
Section~\ref{sec:security} presents the security model.
Section~\ref{sec:performance} discusses optimizations.
Section~\ref{sec:implementation} summarizes implementation and SDKs.
Section~\ref{sec:related} reviews related work. Section~\ref{sec:future}
outlines future directions and Section~\ref{sec:conclusion} concludes.

\section{Design Goals}\label{sec:design-goals}
The design of the Agent Directory Service (ADS) is driven by a set of
interrelated goals intended to balance openness, verifiability, and operational
pragmatism. We articulate these goals along with resulting architectural
constraints.

\paragraph{G1 Extensible Schema Surface.} The core representation \emph{MUST}
support forward-compatible evolution without invalidating previously published
records. OASF extensions (e.g., MCP server descriptors, evaluation metrics,
prompt bundles) \emph{SHOULD} compose additively and remain ignorable by nodes
that do not implement them. This motivates a loosely coupled extension registry
and strongly typed, versioned schema identifiers.

\paragraph{G2 Capability-Centric Discovery.} Discovery performance \emph{MUST}
scale sub-linearly with the total number of records for common capability
queries (skills, domains, features). This necessitates a separation between
capability indices and physical storage location (two-level mapping) and
encourages hierarchical taxonomies to bound combinatorial explosion.

\paragraph{G3 Verifiable Integrity and Provenance.} All records \emph{MUST} be
immutable and addressable by cryptographic digest (CID). Optional signing layers
(Sigstore) \emph{SHOULD} provide provenance without imposing key management
burden on casual publishers. Integrity verification \emph{MUST NOT} require
trusted intermediaries.

\paragraph{G4 Interoperability via Reuse.} The system \emph{SHOULD} reuse
existing, widely deployed distribution substrates (OCI registries, ORAS tooling)
to avoid bespoke artifact transport stacks and to leverage existing caching,
access control, and scanning ecosystems.

\paragraph{G5 Federated Operation.} Distinct organizations \emph{MUST} be able
to operate autonomous registries while participating in a shared discovery
plane. Federation \emph{SHOULD NOT} require global consensus on non-critical
metadata nor impose a single root of trust.

\paragraph{G6 Multi-Dimensional Query Composition.} Queries combining at least
one skill plus optional domain and feature filters \emph{MUST} be executable
through efficient index intersections. Domain- or feature-only queries are
intentionally disallowed to preserve index locality and predictable complexity
bounds.

\paragraph{G7 Elastic Replication Strategies.} Deployments \emph{SHOULD}
flexibly choose between on-demand retrieval, proactive caching, and strategic
replication depending on workload latency and availability requirements. The
architecture \emph{MUST} not hard-code a single replication policy.

\paragraph{G8 Security-in-Depth.} Threat mitigation (supply chain tampering,
spoofing, targeted DoS amplification) \emph{SHOULD} rely on layered primitives:
content addressing, transparent signatures, decentralized lookups, cache
isolation. No single compromised component \emph{SHOULD} yield undetectable data
forgery.

\paragraph{G9 Operational Observability.} Index accuracy, replication lag,
signature verification failure rates, and cache effectiveness \emph{SHOULD} be
observable through exportable metrics to enable adaptive policy tuning (future
work elaborates evaluative telemetry extensions).

\paragraph{Resulting Constraints.}
\begin{itemize}[leftmargin=1.2em]
	\item Mutable metadata (e.g., dynamic performance measurements) are
	represented as new versioned records or referrer objects, never in-place
	edits.
	\item All capability-bearing identifiers map first to CIDs; peer discovery
	is a subsequent step (precluding location-dependent indexing shortcuts).
	\item Extension collision avoidance requires globally unique (namespace,
	name, version) triples.
	\item Discovery latency targets favor bounded taxonomy depth over
	arbitrarily deep ontologies.
\end{itemize}

These goals collectively justify the layered decomposition detailed in
Sections~\ref{sec:architecture} and \ref{sec:storage}.

\section{System Architecture Overview}\label{sec:architecture}
The AGNTCY Agent Directory Service (ADS) is architected as a layered, federated system for scalable agent capability discovery and artifact distribution. Its main components and flows are:

\paragraph{Layered Model.} ADS is organized into the following layers:
\begin{itemize}
	\item \textbf{Schema Layer:} Defines agent records using the Open Agentic Schema Framework (OASF), supporting extensible, versioned metadata for skills, domains, features, and evaluation signals.
	\item \textbf{Indexing Layer:} Maintains capability-centric indices mapping skills, domains, and features to sets of agent record CIDs~\cite{rfc6920}. Indices are published as immutable OCI artifacts and distributed via a DHT.
	\item \textbf{Storage Layer:} Stores agent records and referrers as OCI artifacts in one or more registries, backed by object storage (e.g., S3, GCS, Azure Blob). Records are content-addressed by cryptographic digest (SHA-256)~\cite{rfc6920}.
	\item \textbf{Distribution Layer:} Uses mature OCI registry protocols~\cite{oci_distribution,oci_image} and ORAS tooling for artifact transport, caching, and replication. Federation allows organizations to operate autonomous registries while participating in a shared discovery plane.
	\item \textbf{Security Layer:} Provides cryptographic integrity, provenance, and verifiability via content addressing and Sigstore-based signing~\cite{openid_auth}.
\end{itemize}

\paragraph{Main Flows.}
\begin{enumerate}
	\item \textbf{Publication:} Publishers create agent records (OASF), push them as OCI artifacts to a registry, and announce capability indices to the DHT. Optional referrers (signatures, evaluations) are attached using OCI referrer APIs.
	\item \textbf{Discovery:} Clients query the DHT for capability indices, resolve CIDs to registry endpoints, and retrieve agent records and associated referrers. Multi-dimensional queries intersect skill, domain, and feature indices for precise selection.
	\item \textbf{Replication and Federation:} Registries may replicate artifacts and indices across boundaries for performance, compliance, or availability. Federation is policy-driven and does not require global mirroring.
	\item \textbf{Verification:} Clients verify record integrity using digest addressing~\cite{rfc6920} and, if present, validate signatures and provenance using Sigstore~\cite{openid_auth}.
\end{enumerate}

This architecture enables scalable, secure, and extensible agent discovery and artifact distribution, supporting federated operation, multi-dimensional queries, and future schema evolution.

\section{Storage Architecture}\label{sec:storage}
The storage subsystem provides durable, verifiable retention of agent records
and related referrers while enabling elastic distribution strategies. It centers
on cryptographic Content Identifiers (CIDs) and leverages OCI-compatible
registries for transport and persistence~\cite{oci_distribution,oci_image}.

\subsection*{Time Sequence: Record Publication and Retrieval}
\begin{figure}[h]
	\centering
	\begin{tikzpicture}[x=1.6cm, y=1.3cm, font=\footnotesize]
		\node (pub)    at (0,0)   {\textbf{Publisher}};
		\node (reg)    at (1.7,0)   {\textbf{Registry}};
		\node (dht)    at (3.4,0)   {\textbf{DHT}};
		\node (cli)    at (5.1,0)   {\textbf{Client}};

		\foreach \x in {0,1.7,3.4,5.1} {\draw[dashed,gray] (\x,-0.3) -- (\x,-9.5);}

		\node[anchor=west] at (0,-1.1) {\textit{Publish}};
		\draw[->,thick] (0,-1.5) -- (1.7,-1.5) node[midway,above,sloped] {Push record};
		\draw[->,thick] (0,-2.1) -- (3.4,-2.1) node[midway,above,sloped] {Announce index};
		\draw[->,thick] (0,-2.7) -- (3.4,-2.7) node[midway,below,sloped] {Announce locator};

		\draw[->,thick] (3.4,-3.3) -- (3.4,-4.0) node[midway,right] {Replicate index};

		\node[anchor=west] at (0,-4.7) {\textit{Retrieve}};
		\draw[->,thick] (5.1,-5.2) -- (3.4,-5.2) node[midway,above,sloped] {Query skill};
		\draw[->,thick] (3.4,-5.8) -- (5.1,-5.8) node[midway,below,sloped] {Return CIDs};
		\draw[->,thick] (5.1,-6.4) -- (3.4,-6.4) node[midway,above,sloped] {Query locator};
		\draw[->,thick] (3.4,-7.0) -- (5.1,-7.0) node[midway,below,sloped] {Return registry};
		\draw[->,thick] (5.1,-7.6) -- (1.7,-7.6) node[midway,above,sloped] {Pull record};
		\draw[->,thick] (1.7,-8.2) -- (5.1,-8.2) node[midway,below,sloped] {Return record};

		\draw[->,thick,dashed] (5.1,-8.8) -- (1.7,-8.8) node[midway,above,sloped] {Fetch referrer};
		\draw[->,thick,dashed] (1.7,-9.4) -- (5.1,-9.4) node[midway,below,sloped] {Return referrer};

		\node[anchor=west,font=\footnotesize] at (0,-10.2) {Solid: main flow, Dashed: optional (signatures/evals)};
	\end{tikzpicture}
	\caption{Time sequence diagram: publication and retrieval of an agent record in ADS storage architecture.}
	\label{fig:storage-sequence}
\end{figure}
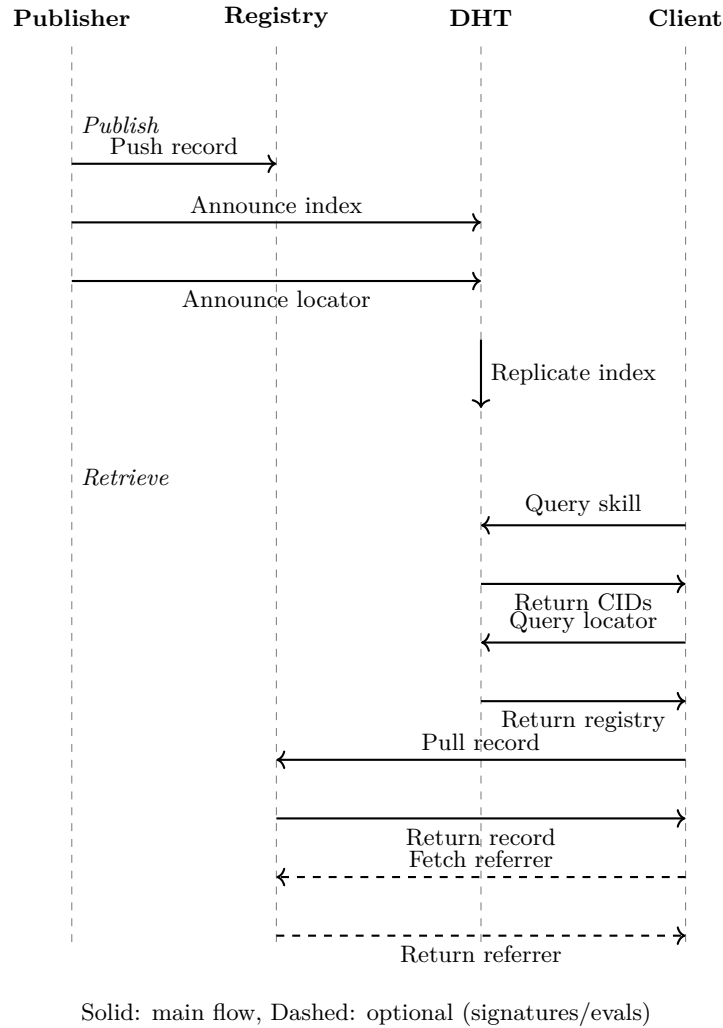

\paragraph{Artifact Layout.} A record is stored as an OCI manifest whose config
media type denotes \texttt{application/vnd.ads.record+json} (illustrative) and
whose layers may include the serialized record plus optional extension payloads.
Tags may group semantically related variants (e.g., capability-sliced views) but
the authoritative identity remains the digest.

\paragraph{Referrers via OCI Index.} OCI referrer APIs (and emerging
registry-native referrer indices) allow attachment of signature bundles,
evaluation datasets, or composition graphs without rewriting the underlying
object.

\paragraph{Transport Reuse Benefits.} By delegating distribution to mature OCI
stacks~\cite{oci_distribution,oci_image}, ADS inherits: (i) parallel layer fetching and caching, (ii) existing CDN
acceleration, (iii) role-based access controls, (iv) integration with
Sigstore-compatible signing flows.

\subsection{Multi-Registry Federation}
Organizations may host distinct registries while participating in a unified
discovery plane. Federation does \emph{not} require global mirroring; only
digests (and optionally minimal metadata) must be resolvable across boundaries.

\paragraph{Federation Modes.}
\begin{itemize}[leftmargin=1.2em]
	\item \textbf{On-Demand Cross-Pull:} Peers retrieve artifacts lazily from
	foreign registries when a CID is discovered.
	\item \textbf{Selective Replication:} Policy-driven mirroring of hot
	artifacts into local registries to reduce latency.
	\item \textbf{Strategic Anchoring:} High-value or compliance-critical
	records replicated across multiple jurisdictions for availability and audit
	resilience.
\end{itemize}

\paragraph{Isolation and Autonomy.} Access policies, retention lifecycles, and
internal replication topologies remain a local concern; only content digests
plus optional provenance metadata traverse administrative boundaries.

\subsection*{Diagram: Storage Layers}
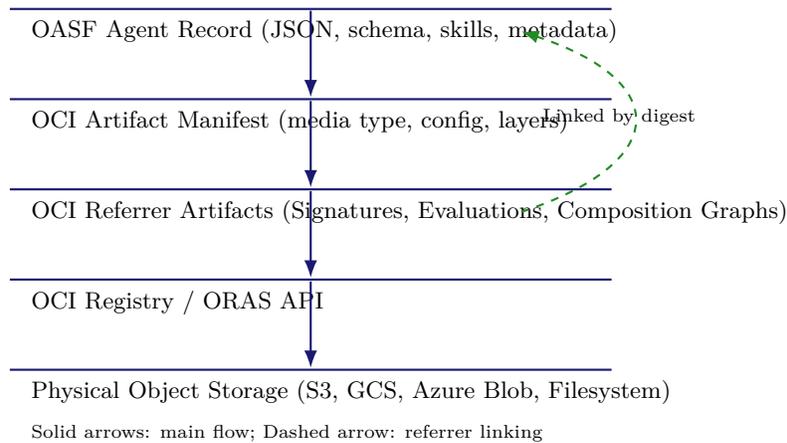
\begin{figure}[h]
	\centering
	\begin{tikzpicture}[font=\small, x=8cm, y=1.2cm]
		\draw[ADSLink, thick] (0,0) -- (1,0);
		\draw[ADSLink, thick] (0,-1) -- (1,-1);
		\draw[ADSLink, thick] (0,-2) -- (1,-2);
		\draw[ADSLink, thick] (0,-3) -- (1,-3);
		\draw[ADSLink, thick] (0,-4) -- (1,-4);

		\node[anchor=west] at (0.02,-0.25) {OASF Agent Record (JSON, schema, skills, metadata)};
		\node[anchor=west] at (0.02,-1.25) {OCI Artifact Manifest (media type, config, layers)};
		\node[anchor=west] at (0.02,-2.25) {OCI Referrer Artifacts (Signatures, Evaluations, Composition Graphs)};
		\node[anchor=west] at (0.02,-3.25) {OCI Registry / ORAS API};
		\node[anchor=west] at (0.02,-4.25) {Physical Object Storage (S3, GCS, Azure Blob, Filesystem)};

		\draw[ADSLink, -Latex, thick] (0.5,-0.02) -- (0.5,-0.98);
		\draw[ADSLink, -Latex, thick] (0.5,-1.02) -- (0.5,-1.98);
		\draw[ADSLink, -Latex, thick] (0.5,-2.02) -- (0.5,-2.98);
		\draw[ADSLink, -Latex, thick] (0.5,-3.02) -- (0.5,-3.98);

		\draw[ADSCite, dashed, -Latex, thick] (0.85,-2.25) .. controls (1.1,-1.7) and (1.1,-0.8) .. (0.85,-0.25);
		\node[font=\scriptsize, anchor=west] at (0.87,-1.2) {Linked by digest};

		\node[font=\scriptsize, anchor=west] at (0.02,-4.7) {Solid arrows: main flow; Dashed arrow: referrer linking};
	\end{tikzpicture}
	\caption{Storage architecture: OASF agent records are stored as OCI manifests in registries backed by object storage. Artifact linking (signatures, evaluations, composition graphs) uses OCI referrers, which reference the manifest by digest.}
	\label{fig:storage-layers}
\end{figure}

\subsection*{Federation Interaction (Textual Schematic)}
\noindent\begin{tabular}{@{}p{2.2cm}p{5.5cm}p{4.3cm}@{}}
	oprule
Component & Role & Example Operations \\
\midrule
Registry A & Origin of skill-tagged NLP / reasoning agents & Push, sign, selective replicate to C \\
Registry B & Vision-focused catalog, caches popular NLP artifacts & On-demand pull from A, replicate subset to C \\
Registry C & Latency-sensitive edge consumers & Pull from A/B, retain strategic replicas \\
Discovery Plane (DHT) & Capability $\rightarrow$ CID and CID $\rightarrow$ Peer mappings & Publish indices, resolve queries \\
\bottomrule
\end{tabular}

\vspace{4pt}
\noindent\textit{Legend:} Cross-registry transfer occurs only after capability
resolution. Replication policy is local; discovery plane never stores bulk
artifact bytes.

\section{Discovery and Taxonomy Model}\label{sec:discovery}
The ADS discovery layer is based on multi-dimensional taxonomies, not just skills but also domains and features, as described in the Internet Draft~\cite{agntcy_ads_draft}. It converts a multi-dimensional capability query into a ranked set of verifiable agent records (CIDs) and, optionally, locates peers or registries that can serve associated artifacts. This section formalizes: (i) taxonomy structure; (ii) index materialization; (iii) the two-level mapping (capability $\rightarrow$ CID, CID $\rightarrow$ peer); (iv) DHT-assisted routing; and (v) query semantics and constraints (Goal G6).

\subsection*{Cross-Server Synchronization and Content Server Selection}
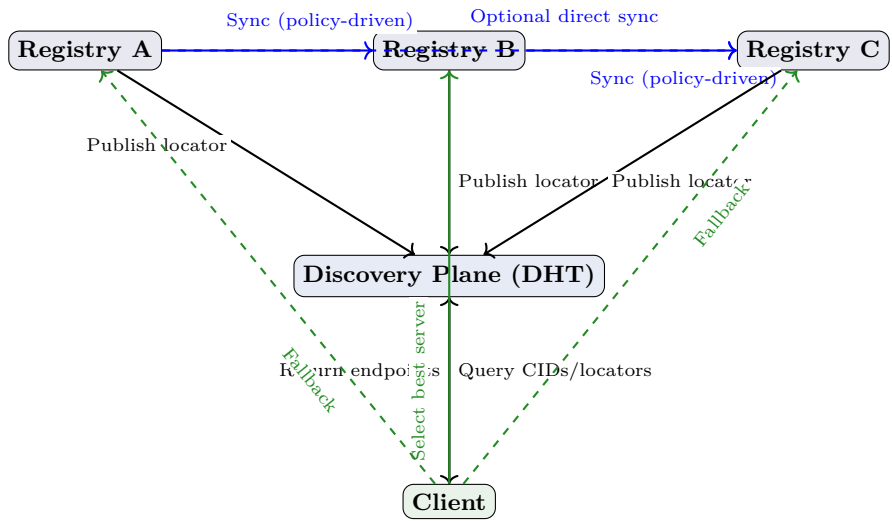
\begin{figure}[h]
	\centering
	\begin{tikzpicture}[x=2.2cm, y=1.5cm, font=\footnotesize]
		\node[draw,rounded corners,fill=ADSLink!10] (regA) at (0,0) {\textbf{Registry A}};
		\node[draw,rounded corners,fill=ADSLink!10] (regB) at (2.2,0) {\textbf{Registry B}};
		\node[draw,rounded corners,fill=ADSLink!10] (regC) at (4.4,0) {\textbf{Registry C}};
		\node[draw,rounded corners,fill=ADSURL!10] (dht) at (2.2,-2) {\textbf{Discovery Plane (DHT)}};
		\node[draw,rounded corners,fill=ADSCite!10] (client) at (2.2,-4) {\textbf{Client}};

	\draw[->,thick,blue] (regA) -- (regB) node[near end,above=6pt,sloped,font=\scriptsize,fill=white,inner sep=1pt] {Sync (policy-driven)};
	\draw[->,thick,blue] (regB) -- (regC) node[near end,below=6pt,sloped,font=\scriptsize,fill=white,inner sep=1pt] {Sync (policy-driven)};
	\draw[->,thick,blue,dashed] (regA) -- (regC) node[pos=0.7,above=8pt,sloped,font=\scriptsize,fill=white,inner sep=1pt] {Optional direct sync};

	\draw[->,thick] (regA) -- (dht) node[pos=0.4,left=2pt,font=\scriptsize,fill=white,inner sep=1pt] {Publish locator};
	\draw[->,thick] (regB) -- (dht) node[pos=0.6,right=2pt,font=\scriptsize,fill=white,inner sep=1pt] {Publish locator};
	\draw[->,thick] (regC) -- (dht) node[pos=0.6,right=2pt,font=\scriptsize,fill=white,inner sep=1pt] {Publish locator};

	\draw[->,thick] (client) -- (dht) node[pos=0.6,right=2pt,font=\scriptsize,fill=white,inner sep=1pt] {Query CIDs/locators};
	\draw[->,thick] (dht) -- (client) node[pos=0.4,left=2pt,font=\scriptsize,fill=white,inner sep=1pt] {Return endpoints};

	\draw[->,thick,ADSCite] (client) -- (regB) node[near start,above=8pt,sloped,font=\scriptsize,fill=white,inner sep=1pt] {Select best server};
	\draw[->,thick,ADSCite,dashed] (client) -- (regA) node[pos=0.3,below=8pt,sloped,font=\scriptsize,fill=white,inner sep=1pt] {Fallback};
	\draw[->,thick,ADSCite,dashed] (client) -- (regC) node[pos=0.7,below=8pt,sloped,font=\scriptsize,fill=white,inner sep=1pt] {Fallback};

	\node[anchor=west,font=\footnotesize] at (0,-5.2) {Blue: sync, Black: DHT publish/query, Green: content selection};
	\end{tikzpicture}
	\caption{Cross-server synchronization and content server selection: registries sync records, publish locators to DHT, and client selects best content server with fallbacks.}
	\label{fig:server-sync-selection}
\end{figure}

\subsection{Hierarchical Skill Taxonomy}\label{subsec:skill-taxonomy}
Skills represent active competencies an agent can perform (e.g.,
	\texttt{nlp.summarization.abstractive},
	\texttt{vision.classification.multilabel}, \texttt{reasoning.chain-of-thought}).
In this paper, we represent taxonomy nodes using a dotted notation to indicate the path in the tree (e.g., \texttt{nlp.summarization.abstractive} is a child of \texttt{nlp.summarization}, which is a child of \texttt{nlp}).
The taxonomy is intentionally \emph{shallow-to-moderate depth} (target depth
$\leq 4$) to balance expressivity and index fan-out. Each skill node has:
\begin{itemize}[leftmargin=1.2em]
	\item A fully qualified identifier (FQID) using dotted path notation.
	\item Optional synonyms (not indexed directly; used for query expansion in
	clients).
	\item A monotonic semantic version for definition evolution (major
	increments ONLY when meaning changes incompatibly).
	\item A parent pointer except for roots.
\end{itemize}
Skill definitions are published as versioned OASF extension documents
addressable by CIDs. Index nodes reference the CID of the skill definition
rather than embedding mutable semantics inline.

\paragraph{Taxonomy Governance.} Multiple federated authorities MAY publish
subtrees under distinct namespace roots (e.g., \texttt{core.nlp},
\texttt{enterprise.compliance}). Clients treat unknown namespaces as opaque
leaves; this preserves forward compatibility (Goal G1).

\subsection{Additional Taxonomies (Domain, Feature)}\label{subsec:other-taxonomies}
Besides skills, ADS supports orthogonal classification axes:
\begin{description}[leftmargin=1.4em, style=nextline]
	\item[Domain] Contextual application scope (e.g., \texttt{healthcare},
	\texttt{finance.retail}, \texttt{legal.contracts}). Domains capture
	regulatory, linguistic, or data-governance context.
	\item[Feature] Optional binary or categorical traits (e.g.,
	\texttt{streaming-output}, \texttt{low-latency},
	\texttt{hardware.accelerated}, \texttt{eval.safety.v1-pass}). These are
	intentionally flatter; deep hierarchies would inflate intersection
	complexity.
\end{description}
Each dimension is indexed independently; multi-dimensional queries intersect
posting lists at the CID layer (Section~\ref{subsec:query-semantics}).

\paragraph{Non-Goals.} ADS does \emph{not} natively index arbitrary free-form
tags. Free-form descriptors can be embedded inside records and surfaced via
external search overlays, preserving core index performance predictability.

\subsection{Two-Level Mapping: Skills\texorpdfstring{$\rightarrow$}{->}CIDs and
CIDs\texorpdfstring{$\rightarrow$}{->}Peers}\label{subsec:two-level}
Discovery is decomposed for scale and resilience:
\begin{enumerate}[leftmargin=1.2em]
	\item Capability-level indices map skill/domain/feature keys to \emph{sets
	of agent record CIDs}. These sets are immutable snapshots partitioned by
	taxonomy node and optionally by time window or version epoch.
	\item A separate resolution layer maps a record CID to one or more serving
	endpoints: (i) OCI registry references (registry host + repository path +
	digest); (ii) optional peer addresses for direct transfer (future); (iii)
	auxiliary mirrors.
\end{enumerate}
Decoupling prevents high-churn location metadata from invalidating large
capability index structures (Goals G2, G7). Capability indices thus remain
stable even as replication or hosting topology evolves.

\paragraph{Index Object Structure.} An index object is itself an OCI artifact
whose payload lists (compressed, canonical order) the CIDs associated with a
taxonomy key. Optional delta objects encode additions/removals referencing a
base index CID to reduce re-publication overhead.

\paragraph{Referrer Usage.} Referrers attach evaluation summaries or trust
annotations to either (a) individual CIDs, or (b) index objects (e.g.,
quality-weighted ordering heuristics) without mutating originals.

\subsection{DHT-Based Query Resolution}\label{subsec:dht-resolution}
The Distributed Hash Table (DHT) stores lightweight locator entries:
\begin{itemize}[leftmargin=1.2em]
	\item Key: Hash(skill FQID) $\rightarrow$ latest index CID(s).
	\item Key: Hash(domain FQID) $\rightarrow$ latest index CID(s).
	\item Key: Hash(feature identifier) $\rightarrow$ latest index CID(s).
	\item Key: CID $\rightarrow$ one or more registry endpoints (host,
	repository, digest) plus optional signature bundle reference.
\end{itemize}
Entries are small (tens to hundreds of bytes) enabling high replication factors
for availability. DHT nodes perform \emph{validation} by checking that index
artifact digests match announced CIDs and optionally verifying embedded
signatures (preview of Section~\ref{sec:security}).

\subsection*{Time Sequence: Cross-Server Synchronization and Content Server Selection}
\begin{figure}[h]
	\centering
	\begin{tikzpicture}[x=1.6cm, y=1.2cm, font=\footnotesize]
		\node (regA) at (0,0) {\textbf{Registry A}};
		\node (regB) at (1.6,0) {\textbf{Registry B}};
		\node (regC) at (3.2,0) {\textbf{Registry C}};
		\node (dht)  at (4.8,0) {\textbf{DHT}};
		\node (cli)  at (6.4,0) {\textbf{Client}};

		\foreach \x in {0,1.6,3.2,4.8,6.4} {\draw[dashed,gray] (\x,-0.3) -- (\x,-8.5);}

		\node[anchor=west] at (0,-1.0) {\textit{Publish/Sync}};
		\draw[->,thick] (0,-1.4) -- (1.6,-1.4) node[midway,above,sloped] {Sync record};
		\draw[->,thick] (1.6,-1.8) -- (3.2,-1.8) node[midway,above,sloped] {Sync record};
		\draw[->,thick,dashed] (0,-2.2) -- (3.2,-2.2) node[midway,below,sloped] {Direct sync (optional)};

		\node[anchor=west] at (0,-2.8) {\textit{Locator Publish}};
		\draw[->,thick] (0,-3.2) -- (4.8,-3.2) node[midway,above,sloped] {Publish locator};
		\draw[->,thick] (1.6,-3.6) -- (4.8,-3.6) node[midway,above,sloped] {Publish locator};
		\draw[->,thick] (3.2,-4.0) -- (4.8,-4.0) node[midway,above,sloped] {Publish locator};

		\node[anchor=west] at (0,-4.6) {\textit{Client Query/Select}};
		\draw[->,thick] (6.4,-5.0) -- (4.8,-5.0) node[midway,above,sloped] {Query locators};
		\draw[->,thick] (4.8,-5.4) -- (6.4,-5.4) node[midway,below,sloped] {Return endpoints};
		\draw[->,thick,ADSCite] (6.4,-6.0) -- (1.6,-6.0) node[midway,above,sloped] {Select best server};
		\draw[->,thick,ADSCite,dashed] (6.4,-6.4) -- (0,-6.4) node[midway,above,sloped] {Fallback};
		\draw[->,thick,ADSCite,dashed] (6.4,-6.8) -- (3.2,-6.8) node[midway,above,sloped] {Fallback};

		\node[anchor=west,font=\footnotesize] at (0,-8.2) {Solid: main flow, Dashed: optional, Green: content selection};
	\end{tikzpicture}
	\caption{Time sequence diagram: cross-server synchronization and content server selection.}
	\label{fig:sync-time-sequence}
\end{figure}
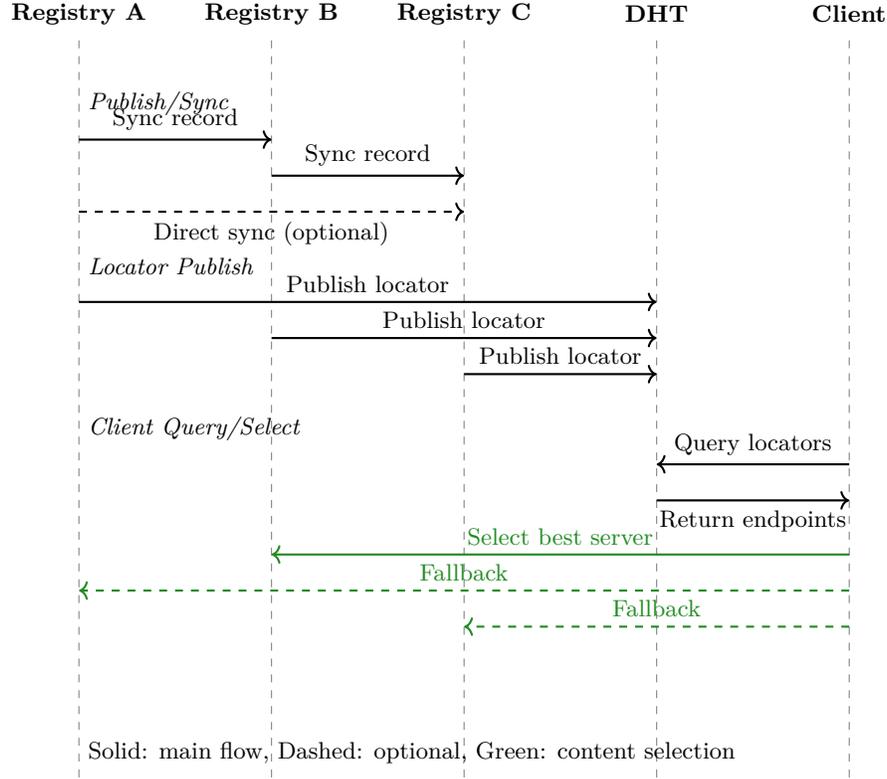

\paragraph{Update Propagation.} Publication flows:
\begin{enumerate}[leftmargin=1.2em]
	\item Publisher pushes new record (OCI manifest) \emph{and} any necessary
	index delta artifact to its registry.
	\item Publisher announces (skill hash $\rightarrow$ index CID) to the DHT.
	\item Peers retrieving the index delta reconstruct full posting list;
	negative deltas (removals) are disallowed—removal is modeled as index
	supersession with a pruned list.
	\item CID $\rightarrow$ location entries are published independently,
	allowing late replication.
\end{enumerate}

\paragraph{Staleness Handling.} Queries tolerate short-lived staleness: if an
index CID resolves but downstream artifact fetch fails (e.g., replication lag),
clients MAY fallback to secondary registry endpoints from the CID locator
mapping.

\subsection{Multi-Dimensional Query Semantics}\label{subsec:query-semantics}
Let a query specify: one required skill $s$, optional domain set $D = \{d_1,
..., d_m\}$, and optional feature set $F = \{f_1, ..., f_n\}$. Resolution
proceeds:
\begin{enumerate}[leftmargin=1.2em]
	\item Resolve $s$ to index CID $I_s$ via DHT; fetch posting list $P_s$.
	\item For each $d_i$, resolve $I_{d_i}$; fetch $P_{d_i}$. Likewise for each $f_j$ yielding $P_{f_j}$.
	\item Compute intersection: $C = P_s \cap (\bigcap_i P_{d_i}) \cap (\bigcap_j P_{f_j})$ using increasing list length ordering to minimize intermediate cardinality.
	\item For each CID in $C$, resolve serving endpoints; optionally fetch referrers to enrich ranking (e.g., filter by evaluation score threshold).
\end{enumerate}
If $D$ or $F$ is empty, they are neutral. Queries omitting a skill are rejected
(Goal G6) to avoid unbounded scanning across broad domain or feature groupings.

\paragraph{Complexity Considerations.} With sorted posting lists, intersection
is $O(\sum_k |P_k|)$ but practical cost is lowered by ordering lists from
smallest to largest cardinality. Encouraging moderately balanced taxonomy
branches reduces pathological skew. Clients MAY cache high-frequency $P_s$
lists, bounded by signature or revision epochs.

\paragraph{Ranking Extensions.} Base order is unspecified. Optional ranking
strategies (future work) can incorporate: (i) recency of record CID publication,
(ii) presence of verified signatures, (iii) evaluation score aggregates, (iv)
declared performance metadata. These are layered post-filters and do not alter
core set semantics.

\paragraph{Failure Modes.} Missing $I_{d_i}$ entries degrade gracefully—lists
are treated as universal (i.e., no further restriction). Corruption (digest
mismatch) triggers client-side rejection and fallback to cached previous index
CID if still within a grace epoch.

\paragraph{Example.} Query: \texttt{skill=nlp.summarization.abstractive,
domain=healthcare, features=streaming-output+eval.safety.v1-pass}. Intersection
retrieves candidate CIDs whose records declare that skill, are tagged under the
healthcare domain, and possess BOTH streaming-output feature and the safety
evaluation feature. If no intersection exists, clients MAY relax (drop longest
list dimension first) to present near-miss suggestions—outside core protocol but
viable at the UX layer.

\subsection{Rationale and Alternatives}\label{subsec:rationale}
Alternative designs considered:
\begin{itemize}[leftmargin=1.2em]
	\item \emph{Single-Level (Skill $\rightarrow$ Peers):} Collapses mapping but
	forces index churn upon replication changes, violating separation of
	concerns.
	\item \emph{Full Text Search Core:} Flexible but sacrifices determinism,
	increases operational complexity (inverted index sharding), and weakens
	verifiability of minimal canonical posting lists.
	\item \emph{Deep Ontologies ($>$6 levels):} Higher expressivity but
	exacerbates sparse branches and large posting list unions when collapsing
	queries across adjacent conceptual layers.
\end{itemize}
The adopted two-level model preserves stable cryptographic sets for capability
semantics while allowing elastic, policy-driven replication beneath.

\section{Security Model}\label{sec:security}
The ADS security model layers immutable content addressing, optional transparent
signing, decentralized index validation, and controlled query semantics to
reduce forgery, spoofing, and tampering risk surfaces while avoiding centralized
trust anchors.

\subsection{Cryptographic Integrity and Hash Addressing}\label{subsec:integrity}
Each agent record, index object, delta, and referrer is serialized
deterministically (Section~\ref{sec:storage}) and addressed by a cryptographic
digest (SHA-256). Integrity guarantees:
\begin{itemize}[leftmargin=1.2em]
	\item \textbf{Collision Resistance:} Adversaries cannot feasibly craft two
	distinct payloads with the same digest.
	\item \textbf{Tamper Evidence:} Any bit-level mutation changes the digest,
	invalidating downstream references (index entries, referrers, signatures).
	\item \textbf{Content-Defined Identity:} Identical logical content always
	yields the same CID, enabling deduplication and cross-registry reuse.
\end{itemize}
Digest verification occurs on retrieval before higher-layer processing (ranking,
evaluation filtering). Clients \emph{MUST} reject mismatched digests.

\paragraph{Canonicalization Enforcement.} Toolchains SHOULD embed a
canonicalization check (hash-of-hash self attest or manifest annotation)
allowing validators to detect non-canonical encodings that could induce
duplicate semantic objects.

\subsection{Provenance via Sigstore}\label{subsec:provenance}
Provenance is optional yet recommended for supply chain verifiability. Sigstore
provides:
\begin{enumerate}[leftmargin=1.2em]
	\item \emph{Keyless Signing:} Short-lived certificates bound to an OIDC
	identity (e.g., organization SSO) reduce private key custody risk.
	\item \emph{Transparency Log Inclusion:} Signatures (record digest +
	metadata) are recorded in an append-only, auditable log; clients can require
	inclusion proof.
	\item \emph{Rekor Integration for Referrers:} Signature bundles are stored
	as referrer artifacts referencing the target CID, avoiding content mutation.
\end{enumerate}
Verification workflow:
\begin{enumerate}[leftmargin=1.2em]
	\item Retrieve record (digest $d$) and associated signature referrer
	(contains certificate chain + Rekor entry digest).
	\item Validate certificate expiration and issuer chain; check SAN matches
	expected publisher pattern (policy-defined).
	\item Verify signature over $d$; query transparency log for inclusion proof;
	optionally enforce Signed Certificate Timestamp (SCT) presence if log
	supports real-time gossip.
\end{enumerate}
Policies MAY express: (i) required issuer(s); (ii) minimum transparency
inclusion depth (blocks); (iii) revocation strategies (treat future-dated certs
or time skew anomalies as policy violations).

\paragraph{Detached vs Embedded Signatures.} ADS favors detached referrer
signatures to preserve the immutability of the core record object and allow
multiple independent signers (e.g., publisher, auditor, evaluator) without
combinatorial record rewriting.

\subsection{Trust Boundaries and Isolation}\label{subsec:trust}
ADS distinguishes between:\\
\begin{tabular}{@{}p{3.2cm}p{9.5cm}@{}}
	\textbf{Publisher Trust} & Entity asserting a record's authenticity;
	validated through Sigstore identity or out-of-band trust anchors. \\
	\textbf{Registry Trust} & Operational durability and availability provider;
	not implicitly trusted for data integrity (digest verification required). \\
	\textbf{Discovery Plane Trust} & DHT peers assumed
	\emph{honest-but-curious}; malicious entries detectable via digest mismatch
	or signature absence. \\
	\textbf{Client Policy Trust} & Local enforcement of acceptable identities,
	evaluation thresholds, required signature presence. \\
\end{tabular}

\paragraph{Separation of Concerns.} Compromise of any single boundary (e.g.,
registry) cannot forge a valid signed record without corresponding keys; forging
an unsigned record is detectable via digest mismatch against previously cached
indices.

\subsection{Threat Mitigation and Attack Surfaces}\label{subsec:threats}
\begin{description}[leftmargin=1.4em, style=nextline]
	\item[Artifact Tampering] Mitigated by digest addressing + optional signatures; detection via hash verify failure.
	\item[Signature Forgery] Prevented by short-lived cert issuance + transparency auditing; clients can require log inclusion.
	\item[Downgrade / Substitution] Old or unrelated records cannot masquerade as targets due to digest mismatch; index supersession uses new CIDs.
	\item[Sybil Index Poisoning] Large-scale fake index announcements limited by requirement to host corresponding artifact (fetch+verify) and potential stake / rate-limiting in future DHT hardening.
	\item[Replay of Stale Indices] Clients MAY enforce epoch or monotonic sequence fields inside index metadata to reject unexpectedly older CIDs.
	\item[Targeted DoS Amplification] Capability queries require a skill dimension, bounding worst-case result set size relative to high-cardinality free-form tag floods.
	\item[Metadata Inference Leakage] Minimal index payloads (CIDs only) reduce leakage; optional encrypted referrer layers could cloak sensitive evaluation metrics (future work).
\end{description}

\paragraph{Defense-in-Depth Summary.} Table~\ref{tab:security-layers} (future)

\begin{table}[h]
	\centering
	\caption{Defense-in-Depth: Security Layers and Controls}
	\label{tab:security-layers}
	\begin{tabular}{@{}p{4.5cm}p{8cm}@{}}
		Threat & Layered Controls \\
		\midrule
		Artifact Tampering & Content addressing, signatures \\
		Signature Forgery & Cert issuance, transparency logs \\
		Downgrade/Substitution & Digest matching, index supersession \\
		Sybil Index Poisoning & Artifact hosting, rate-limiting \\
		Replay of Stale Indices & Epoch/sequence enforcement \\
		Targeted DoS Amplification & Skill dimension bounding \\
		Metadata Inference Leakage & Minimal index payloads, encrypted referrers \\
		\bottomrule
	\end{tabular}
\end{table}
will map threats to layered controls (content addressing, signing, transparency,
intersection semantics, replication diversity).

\section{Performance and Replication}\label{sec:performance}
Performance characteristics of ADS emerge from: (i) the separation of capability
indices from artifact storage; (ii) the bounded taxonomy depth; (iii) selective
replication and caching policies; and (iv) immutable, deduplicated objects
enabling widespread CDN and registry reuse. This section formalizes the
principal strategies and trade-offs.

\subsection{On-Demand Retrieval}
On-demand (lazy) retrieval defers artifact transfer until a query result is
actually materialized by a client. Workflow:
\begin{enumerate}[leftmargin=1.2em]
	\item Query executes index intersections; only posting lists and
	(optionally) small ranking referrers are fetched.
	\item Client selects a subset (e.g., top \texttt{k}) of candidate CIDs based
	on local policy (signature presence, evaluation score, freshness).
	\item Artifact manifests and required layers are pulled from the origin
	registry (or nearest mirror discovered via CID locator entries).
\end{enumerate}
Advantages: zero wasted bandwidth for unmaterialized candidates; simplified
cache invalidation (immutable CIDs). Disadvantages: elevated tail latency for
first access to cold artifacts. Tail latency can be partially amortized by
parallel prefetch of manifests (lightweight) while deferring heavy layer blobs
until user intention is confirmed.

\paragraph{Latency Bound.} Let $T_q$ be query intersection time (typically
dominated by network round trips for DHT index CID resolution) and $T_f$ the
first-byte fetch time for a cold artifact. On-demand adds $T_q + T_f$ to
time-to-usable metadata. With regional registry mirrors, $T_f$ is minimized;
absent mirrors, cross-region RTTs dominate. Caching layers
(Section~\ref{subsec:proactive-caching}) target reducing $T_f$ without incurring
full replication costs.

\subsection{Proactive Caching}\label{subsec:proactive-caching}
Proactive caching retains frequently requested posting lists (capability index
payloads) and optionally small artifact manifests near query execution points
(e.g., in the Query API or edge function layer). Policies:
\begin{itemize}[leftmargin=1.2em]
	\item \textbf{Frequency-Based:} Cache posting lists whose access count over
	a sliding window exceeds a threshold.
	\item \textbf{Recency + Cost:} Prioritize caching of lists with large
	cardinality reduction effect (i.e., (skill, domain, feature) nodes that
	sharply constrain intersections).
	\item \textbf{Hybrid Significance Score:} $S = \alpha f + \beta r + \gamma
	c$ where $f$ is normalized frequency, $r$ is recency weight, $c$ is
	compression ratio improvement (benefit of avoiding another fetch). Tunable
	coefficients $(\alpha,\beta,\gamma)$ allow environment-specific tuning.
\end{itemize}
Cached lists include a weak ETag (hash of concatenated CID sequence); if a DHT
announcement yields the same ETag, revalidation is skipped. Because indices are
immutable, stale eviction is driven by superseding index CIDs—not time-based
expiry—reducing validation chatter.

\paragraph{Impact.} Caching primarily reduces $T_q$ (query intersection
latency). When $P_s$ (skill posting list) constitutes \(>60\%\) of intersection
bytes, caching yields near-linear reduction in total query time for repeated
workloads with stable taxonomy usage patterns.

\subsection{Strategic Replication}
Strategic replication involves copying full artifact blobs (not just indices) to
additional registries or edge caches based on policy drivers:
\begin{description}[leftmargin=1.4em, style=nextline]
	\item[Hotset Replication] Top $N$ CIDs by query appearance frequency or
	post-selection usage. Periodically recomputed (e.g., hourly) to adjust to
	shifting workload distributions.
	\item[Skill-Critical Replication] Mandatory replication for designated
	high-criticality skills (e.g., compliance validators, safety screening
	agents) to guarantee local availability for governance pipelines.
	\item[Jurisdictional Anchoring] Replicate artifacts across distinct legal
	regions (e.g., EU, US, APAC) to mitigate data residency or regulatory
	downtime concerns.
	\item[Evaluation Cohort Pinning] Replicate all agents sharing a benchmark
	cohort to enable low-latency comparative evaluation runs.
\end{description}
Replication events publish new CID locator entries (CID $\rightarrow$ endpoints)
in the DHT without modifying capability indices. Clients may rank endpoints
using a latency heuristic (measured RTT, geo proximity) or historical success
rate.

\paragraph{Cost Model.} Let $B$ be average artifact blob size, $F$ query
frequency for its CID, $L_o$ average latency from origin, $L_r$ latency from
replica, and $C_t$ transfer cost per byte. Replication is beneficial when $F
(L_o - L_r) > k B C_t$ where $k$ encodes acceptable amortization horizon (e.g.,
target break-even over $k$ queries). Operators can approximate $F$ via sampled
Query API logs (excluding PII) aggregated into anonymized frequency sketches.

\subsection{Bandwidth and Latency Optimizations}
Optimization techniques minimize network overhead while preserving
verifiability:
\begin{itemize}[leftmargin=1.2em]
	\item \textbf{Delta Indices:} Large capability posting lists rarely change
	wholesale; publishing additive delta artifacts referencing a base index
	reduces transfer size. Clients materialize full lists incrementally.
	\item \textbf{Bloom Prefilters:} Optional Bloom filters (or XOR filters)
	embedded in index metadata can accelerate early elimination in
	multi-dimensional intersections, reducing full list fetches when
	intersection is predicted empty. False positive rate is tuned (e.g., 1\%) to
	bound wasted fetch attempts.
	\item \textbf{Manifests-First Prefetch:} Parallel retrieval of manifests
	(small JSON) for candidate CIDs while ranking referrers are fetched
	amortizes round trips prior to user selection.
	\item \textbf{Adaptive List Ordering:} Intersection begins with the smallest
	fetched posting list; if size estimates (cardinality hints) reveal a
	different unfetched list is likely smaller, client reorders fetch sequence
	to minimize early intermediate set size.
	\item \textbf{Compression and Canonical Ordering:} Posting lists are
	distributed as sorted, delta-encoded CID digests (variable-length integer
	offsets) achieving high compression ratios (often >5:1) and enabling
	streaming intersection.
	\item \textbf{Client-Side TTL for Failures:} Negative caching (short TTL)
	for unreachable endpoints avoids repeated expensive origin attempts during
	transient outages.
\end{itemize}

\paragraph{End-to-End Latency Decomposition.} A representative cold-path
latency: $T = T_{dht} + T_{idx} + T_{int} + T_{rank} + T_{art}$ where $T_{dht}$
(parallel key lookups), $T_{idx}$ (fetch + decompress posting lists), $T_{int}$
(set intersections), $T_{rank}$ (optional referrer evaluation), $T_{art}$
(artifact manifest fetch). Warm-path removes most of $T_{idx}$ and reduces
$T_{dht}$ to conditional validation checks.

\paragraph{Resilience Considerations.} Because capability indices and blob
replication are decoupled, transient registry unavailability only affects
$T_{art}$ for affected CIDs; queries still resolve candidate sets, enabling
fallbacks or deferred materialization.

\subsection{Trade-offs and Limitations}
Aggressive replication inflates storage and egress costs; insufficient
replication increases $T_f$ for long-tail artifacts. Bloom filters add CPU
overhead and memory footprint; disabled in ultra-latency-sensitive minimal
deployments. Delta indices complicate validation logic (must ensure correct
application order); periodic full snapshot indices mitigate drift. Negative
caching risks masking quick recoveries if TTL too long. Parameter tuning (filter
false positive rate, replication hotset size, cache admission thresholds)
remains deployment-specific.

\paragraph{Summary.} ADS performance strategy prioritizes: (i) immutable,
compressible index artifacts for fast capability set operations; (ii) lazy
artifact transfer for bandwidth efficiency; (iii) adaptive, policy-driven
replication to cap tail latency; and (iv) optional probabilistic prefilters and
compression to bound intersection costs under multi-dimensional queries.

\section{Implementation Details}\label{sec:implementation}
The reference implementation emphasizes minimality, reuse of commodity
components, and clear separation of pure data structures from operational
orchestration. This section outlines core services, SDK abstractions, signing
workflow specifics, and representative schema extensions.

\subsection{Reference Services and Components}
The implementation composes loosely coupled services (each independently
deployable):
\begin{description}[leftmargin=1.4em, style=nextline]
	\item[Index Publisher] Accepts validated agent record submissions, produces
	canonical serialization, pushes OCI artifacts (record + referrers), emits
	index delta artifacts, and announces DHT updates.
	\item[DHT Node] Implements Kademlia-like routing table maintenance,
	key/value storage for capability and CID locator entries, digest validation
	hooks, and optional signature presence checks.
	\item[Registry (External)] Any OCI-compliant registry (e.g., Harbor, GHCR,
	OCI Distribution) storing manifests, layers, and referrers; no ADS-specific
	code required.
	\item[Query API] Stateless HTTP/gRPC endpoint providing multi-dimensional
	query handling, intersection execution, caching of hot posting lists, and
	ranking plugin pipeline.
	\item[Evaluation Ingestor] Consumes evaluation result feeds (benchmark
	harnesses, governance scanners), emits referrer artifacts attaching metrics
	and quality signals to CIDs.
	\item[Signer / Transparency Client] Wraps Sigstore keyless flow, obtains
	ephemeral certs, submits Rekor entries, and publishes detached signature
	bundles as referrers.
\end{description}
Deployment patterns range from a consolidated single-process developer mode
(embedded DHT + lightweight registry proxy) to a production multi-node cluster
with horizontal scaling on Query API and DHT nodes.

\subsection{SDK Interfaces (Python / JavaScript)}
SDKs expose symmetrical primitives across languages:
\begin{itemize}[leftmargin=1.2em]
	\item \texttt{publish(record: AgentRecord, extensions: [...]) -> CID}
	\item \texttt{add\_referrer(target: CID, ref: ReferrerObject) -> CID}
	\item \texttt{query(skill: str, domains: [str] = [], features: [str] = [], limit: int = 50) -> [ResolvedAgent]}
	\item \texttt{verify(cid: CID, require\_signature: bool = false) -> VerificationReport}
	\item \texttt{get\_history(cid: CID) -> [ReferrerDescriptor]} (evaluation snapshots, composition graphs)
\end{itemize}
Representative Python type sketch:
\begin{verbatim}
class AgentRecord(BaseModel):
		schema: str          # OASF schema ID
		version: str         # semantic version
		skills: list[str]
		domains: list[str]
		features: list[str]
		resources: dict      # endpoints, model URIs, MCP server URLs
		metadata: dict       # arbitrary extension-friendly map

class ReferrerObject(BaseModel):
		kind: Literal['signature','evaluation','composition','annotation']
		target_cid: str
		payload: dict        # structure depends on kind
		created_at: datetime

class VerificationReport(BaseModel):
		cid: str
		digest_match: bool
		signatures: list[dict]   # issuer, subject, log_included
		evaluation_summary: dict # aggregated metrics (optional)
		policy_pass: bool
\end{verbatim}
The JS/TypeScript variant mirrors these contracts for front-end or edge runtime
integration.

\paragraph{Caching Hints.} SDK query responses MAY include weak ETags (hash of
concatenated index CIDs) enabling conditional re-validation to avoid fetching
unchanged posting lists.

\subsection{Signing and Verification Workflow}
End-to-end pipeline for a published record:
\begin{enumerate}[leftmargin=1.2em]
	\item Developer invokes \texttt{publish()} locally; tooling canonicalizes
	and computes digest $d$.
	\item Digest $d$ and manifest are pushed to the registry; index delta
	artifact references $d$ is pushed if new skill/domain/feature associations
	exist.
	\item Tooling initiates Sigstore keyless signing: obtains ephemeral cert,
	signs $d$, submits to transparency log, retrieves log entry (UUID +
	inclusion proof).
	\item Detached signature bundle (certificate + signature + log metadata)
	serialized as referrer OCI artifact (media type
	\texttt{application/vnd.ads.sigbundle+json}).
	\item DHT announcements: (skill hash $\rightarrow$ index CID), (CID
	$\rightarrow$ registry endpoints), optionally (CID $\rightarrow$
	signature-referrer CID).
	\item Client query resolves $d$, fetches record, verifies digest, optionally
	loads signature bundle and validates certificate + log inclusion.
\end{enumerate}
Failure at any step (e.g., signature log unavailability) may still allow
unsigned publication; client-side policy dictates acceptance.

\paragraph{Policy Engine.} A pluggable evaluation applies predicates: minimum
signers, approved OIDC issuers, evaluation metric thresholds, maximum record
age, required feature flags.

\subsection{Schema Extensions (MCP, A2A, Prompts)}
Extensions define structured payload fragments referenced from the core
\texttt{metadata} map or stored as separate layers:
\begin{description}[leftmargin=1.4em, style=nextline]
	\item[MCP Server Descriptor] Declares \texttt{capabilities}, \texttt{tools},
	\texttt{transport} endpoint, aligning with MCP registry schema
	\cite{modelcontextprotocolregistry}.
	\item[A2A Endpoint Metadata] Specifies agent-to-agent protocol handshake
	versions, supported message codecs, and QoS hints referencing
	\cite{a2aspecification}.
	\item[Prompt Bundle] Versioned collection of prompt templates (system, task,
	evaluation) with hashing for each template enabling selective reuse.
	\item[Evaluation Metrics] Structured vector (e.g., latency p95, factuality
	score, safety pass ratio) enabling future machine-readable ranking
	strategies.
	\item[Composition Graph] DAG describing multi-agent orchestration or tool
	chaining (nodes reference constituent agent CIDs) enabling reproducible
	compound behaviors.
\end{description}
Each extension is assigned a globally unique media type and schema URL. Unknown
extensions are safely ignored by baseline clients, preserving forward
compatibility (Goal G1).

\section{Related Work}\label{sec:related}
Efforts to structure discovery and interoperability across emerging AI agent
ecosystems span naming systems, interoperability protocols, identity primitives,
and artifact distribution substrates. We situate ADS with respect to these
dimensions.

\paragraph{Naming and Index Resolution.} The NANDA corpus
~\cite{raskar2025dnsunlockinginternetai}
~\cite{nanda2025upgrade_or_switch_arxiv}
~\cite{nanda2025adaptive_resolver}
~\cite{nanda2025enterprise_perspective}
~\cite{nanda2025registry_survey}
argues for DNS-inspired hierarchical naming, adaptive resolution strategies, and
enterprise governance layers for an ``Internet of AI Agents.'' ADS is
complementary: it does not attempt to standardize mnemonic names or resolver
behaviors, instead treating any external name or handle as an input that
ultimately resolves (directly or indirectly) to capability-bearing,
content-addressed records (CIDs). Where NANDA focuses on resolver adaptability
and governance surface, ADS emphasizes capability taxonomy intersection and
cryptographic object immutability. A future interoperability path envisions name
resolution (NANDA) yielding a capability query or CID seed refined via ADS's
multi-dimensional indices.

\paragraph{Interoperability Protocol Registries.} The Model Context Protocol
registry \cite{modelcontextprotocolregistry} catalogs MCP servers and tools but
does not define a multi-dimensional capability taxonomy nor a distributed
two-level mapping separating capability indices from storage peers. The A2A
initiative \cite{a2aspecification} introduces protocol-level interoperability
for agent communication; ADS instead targets discovery semantics independent of
runtime messaging protocols. Proposed blockchain-oriented approaches (e.g.,
ERC-8004 trustless agents \cite{erc8004}) explore on-chain attestations and
economic primitives; ADS reuses off-chain content distribution for performance
and cost efficiency while enabling cryptographic verification through signatures
and digests.

\paragraph{Content Distribution and Artifact Reuse.} OCI distribution
\cite{oci_distribution_spec} and ORAS tooling have demonstrated robust
scalability for immutable artifact delivery in container and artifact
ecosystems. ADS directly reuses these semantics (push/pull/referrers, digest
addressing) rather than introducing a bespoke transport, enabling transparent
adoption of registry caching, access policies, and scanning workflows. This
differs from proposals centered purely on hierarchical name delegation that do
not intrinsically bind names to verifiable digests.

\paragraph{Decentralized Routing and Lookup.} DHT-based routing (e.g., Kademlia
deployments underpinning IPFS \cite{ipfs_kademlia}) informs ADS's design for
distributing responsibility of capability-to-CID and CID-to-peer mappings.
Unlike generic content routing, ADS constrains query structure (must include at
least one skill dimension) to preserve bounded intersection complexity (Goal G6)
while allowing federation without global mirroring.

\paragraph{Identity, Integrity, and Provenance.} Identity-focused initiatives
(Microsoft Entra Agent ID \cite{simons2025agentid}) address enterprise lifecycle
management, credential binding, and governance. ADS focuses on artifact-level
immutability (CIDs) and optional signing (Sigstore integration, forthcoming
section) rather than operator directory management. These layers can compose:
organizational identity services can publish or sign ADS records; ADS does not
prescribe identity issuance mechanisms.

\paragraph{Schema Frameworks and Extensibility.} The Open Agentic Schema
Framework (OASF) \cite{oasf} provides a structured, versioned extension surface
analogous in spirit to security telemetry unification efforts (e.g., OCSF
\cite{ocsf}). ADS leverages this to keep the core record minimal while enabling
additive enrichment (evaluations, MCP descriptors, prompt bundles) without
forking the base schema.

\paragraph{Comparative Focus.} Summarizing: (i) naming/index proposals (NANDA)
optimize human-friendly and governance-aware resolution; (ii) interoperability
protocols (MCP, A2A) facilitate runtime context and message exchange; (iii)
identity systems manage agent lifecycle credentials; (iv) on-chain proposals
introduce economic/trust anchoring; (v) ADS centers on scalable,
capability-centric, content-addressed discovery with federated, storage-agnostic
indices. These strata are complementary and layered rather than mutually
exclusive.

\paragraph{Comparison: Semantic Description of Agent Capabilities.}
Table~\ref{tab:comparison} compares MCP, A2A, and NANDA with respect to their support for semantic description of AI agent capabilities (skills, taxonomies, extensibility, and related axes):

\begin{table}[h]
	\centering
	\scriptsize
	\renewcommand{\arraystretch}{1.15}
	\begin{tabularx}{\textwidth}{@{}l>{\raggedright\arraybackslash}X>{\raggedright\arraybackslash}X>{\raggedright\arraybackslash}X>{\raggedright\arraybackslash}X@{}}
			oprule
			extbf{System} & \textbf{Skill/Capability Semantics} & \textbf{Taxonomy Support} & \textbf{Extensibility} & \textbf{Notes} \\
		\midrule
		MCP & Structured (capabilities, tools, endpoints) & Flat or ad-hoc lists & Moderate (schema extensions) & Focus on runtime context, not deep skill taxonomy~\cite{modelcontextprotocolregistry} \\
		A2A & Protocol-level (supported messages, codecs) & Minimal (feature flags) & Limited & Emphasizes agent comms, not capability semantics~\cite{a2aspecification} \\
		NANDA & Hierarchical (DNS-inspired names, skills) & Strong (multi-level, governance-aware) & High (adaptive, federated) & Focus on naming, resolver adaptability, and governance~\cite{raskar2025dnsunlockinginternetai}~\cite{nanda2025upgrade_or_switch_arxiv} \\
		ADS & Structured (OASF, skills, domains, features) & Hierarchical, multi-dimensional & High (schema, referrers, extensions) & Capability-centric, content-addressed, federated indices \\
		\bottomrule
	\end{tabularx}
	\caption{Comparison of MCP, A2A, NANDA, and ADS on semantic description of agent capabilities.}
	\label{tab:comparison}
\end{table}

\section{Future Work}\label{sec:future}
ADS opens several avenues for future research and development:
\begin{itemize}
	\item \textbf{Interoperability with Naming Systems:} Integrating ADS with DNS-inspired agent naming and resolution (e.g., NANDA) to enable seamless capability-centric and mnemonic queries.
	\item \textbf{Advanced Ranking and Evaluation:} Developing richer evaluation metrics, ranking strategies, and trust signals for agent selection, including machine-readable benchmarks and governance overlays.
	\item \textbf{Privacy and Confidentiality Extensions:} Supporting encrypted referrer layers and privacy-preserving evaluation sharing for sensitive agent metadata.
	\item \textbf{Federated Governance:} Exploring policy-driven federation, cross-domain trust anchors, and compliance mechanisms for multi-jurisdictional deployments.
	\item \textbf{Performance Optimization:} Implementing adaptive caching, replication, and query acceleration techniques for large-scale, low-latency agent discovery.
	\item \textbf{Schema Evolution:} Formalizing extension registries and versioning strategies for OASF and related schema surfaces to support emerging agent modalities.
	\item \textbf{Open Source Tooling:} Expanding SDKs, reference implementations, and developer tooling for broader adoption and community-driven innovation.
\end{itemize}

\section{Conclusion}\label{sec:conclusion}
The AGNTCY Agent Directory Service (ADS) provides a scalable, capability-centric, and content-addressed foundation for agent discovery in multi-agent systems. By leveraging the Open Agentic Schema Framework (OASF), OCI registry infrastructure, and cryptographic signing, ADS enables efficient, verifiable, and extensible indexing of agent capabilities, domains, and features. Its layered architecture decouples capability indices from artifact storage, supporting federated operation, adaptive replication, and multi-dimensional queries. ADS complements parallel efforts in agent naming, interoperability, and identity, and is designed for forward compatibility and integration with emerging standards. Ongoing work will further enhance interoperability, ranking, privacy, and performance, positioning ADS as a foundational component for the next generation of agentic ecosystems.

\bibliographystyle{plain}
\bibliography{references}

\end{document}